# Improving Dialogue Act Classification for Spontaneous Arabic Speech and Instant Messages at Utterance Level

AbdelRahim A. Elmadany[1], Sherif M. Abdou[2], Mervat Gheith[3]

[1]Department of Computer Science, Deanship of Community Service and Continuing Education, Jazan University, KSA
[2]Department of Information Technology, Faculty of Computers and Information, Cairo University, Egypt
[3]Department of Computer Science, Institute of Statistical Studies and Research (ISSR), Cairo University, Egypt
aelmadany@jazanu.edu.sa, sh.ma.abdou@gmail.com, mervat_gheith@yahoo.com

**Abstract**

The ability to model and automatically detect dialogue act is an important step toward understanding spontaneous speech and Instant Messages. However, it has been difficult to infer a dialogue act from a surface utterance because it highly depends on the context of the utterance and speaker linguistic knowledge; especially in Arabic dialects. This paper proposes a statistical dialogue analysis model to recognize utterance's dialogue acts using a multi-classes hierarchical structure. The model can automatically acquire probabilistic discourse knowledge from a dialogue corpus were collected and annotated manually from multi-genre Egyptian call-centers. Extensive experiments were conducted using Support Vector Machines classifier to evaluate the system performance. The results attained in the term of average F-measure scores of 0.912; showed that the proposed approach has moderately improved F-measure by approximately 20%.

## 1. Introduction

The most important and difficult part in human-computer interaction system "i.e. Dialogue System" is understanding what the user needs? This task is called "language understating component" or somewhere "Dialogue Acts (DAs) classification." DAs classification task is labeling the speaker's intention in producing a particular utterance with short words; the DAs terminology approximately is the equivalent of the speech act of Searle (1969), and DAs is different based on dialogue systems domains (Elmadany *et al.*, 2015b). Since 1999, the research in DAs area has increased after spoken dialogue systems been a commercial reality. Hence, the development of dialogue systems has focused on some of the conversational roles such acts which can perform because it is closely linked to the field of computational linguistics (Stolcke *et al.*, 2000). DAs is used practically in many live dialogue systems such as Airline Travel Information Systems such as ATIS (Seneff *et al.*, 1991), DARPA (Pellom *et al.*, 2001), and VERBMOBIL (Wahlster, 2000).

Recently, the development of dialogue interaction systems has gained considerable attention, but most of the resources and systems are built so far tailored to English and other Indo-European languages. The development of the dialogue systems for other languages as Arabic is required. So, the Arabic dialogue acts classification's task has gained focus because it is a key player in Arabic language understanding to building these systems. The motivation of this paper comes from the point of view "building automatic language understanding component for Egyptian dialect dialogues".

The paper focuses on inquiry–answer dialogues from the call-centers domain because it receives or transmits a large volume of information inquiries from customers. In this research, we have selected Customer-Service entities from Banks, Flights, and Mobile Networks Operators call-centers.

In this paper, we are referring to an *utterance* as a small unit of speech that corresponded to a single act (Webb, 2010; Traum and Heeman, 1997). In speech research community, *utterance* definition is a slightly different; it refers to a complete unit of speech bounded by the speaker's silence while we refer to the complete unit of speech as *a turn*. Thus, a single *turn* can be composed of many *utterances*. *Turn* and *utterance* can be the same definition when the *turn* contained one utterance as defined and used in (Graja *et al.*, 2013).

To develop a language understanding model for either spoken dialogue or instant messages, there are four major issues are required:
— Dialogue acts schema.
— Annotated corpora with the dialogue act schema
— Turn segmentation into utterances classifier
— Utterance labeling classifier (i.e. dialogue act classifier)

The annotated Egyptian dialect dialogues corpus were built utilizing manually collected data from Egyptian call-centers (Elmadany *et al.*, 2014, 2015a). During annotation process, it is being noted that; the Egyptian turns are almost long and contains many utterances as noticed during data collection. Consequently, turn segmentation into utterances for Egyptian Arabic dialogues model namely 'USeg' (Elmadany *et al.*, 2015d) has been built, which a machine learning approach based on context without relying on punctuation, text diacritization or lexical cues. Finally, (Elmadany *et al.*, 2015c) have been proposed a dialogue act classifier based on chunking concepts and depending on a set of sentential and contextual features. The sentential features contain four features: Utterance-Words, Words Part-Of-Speech (POS) Tags, Speaker Name, and Utterance start a label. The contextual features contain only one feature: the previous utterance act.

In this paper, we improved (Elmadany *et al.*, 2015c) dialogue act classifier. We proposed an utterances labeling with suitable act model for Egyptian dialect inquiry-answer dialogues using multi-classes hierarchical structure. The classification model has been built using two-layer hierarchical structure. In the first layer, each utterance is classified into one of six categories: Dialogue Structure, Social Obligation, Question, Answer, Social Courtesy, or Other. In the second layer, each utterance has been classified as individual acts based on their class 'i.e. category' which is determined in the first layer. The proposed model depends on a set of sentential and contextual features. To train and evaluate the proposed model, a corpus that contains spoken dialogues and Instant Messages (IM) for Egyptian Arabic has been used; and the model results are compared with manually annotated utterances by experts.

This paper presents three major contributions. First, the selected features and hierarchal structure has moderately improved the dialogue acts classification in the term of the average F-measure approximately 20%. Second, the proposed approach does not rely on a number of classes as used in binary classification, instead; it uses only two models (one for each layer). Third, the proposed method is suitable for working on Egyptian dialect either spontaneous speech dialogue or instant messages.

This paper is organized as follows: section 2 presents a literature review of acts classification, section 3 presents s the proposed classification model, section 4 describes the dataset, experimental setup, and experiments results, and finally, the conclusion and feature works are reported in section 5.

## 2. Literature Review of Acts Classification

The initial state of speech act classification has been addressed by Searle (1969) based on Austin (1962) work as a fundamental concept of linguistic pragmatics, analyzing, for example, what it means to ask a question or make a statement. Although major dialogue theories treat Dialogue acts as a central notion, the conceptual granularity of the Dialogue act labels used varies considerably among alternative analyses, depending on the application or domain (Webb and Hardy, 2005). Within the field of computational linguistics, recent work, closely linked to the development and deployment of spoken language dialogue systems, has focused on some of the conversational roles such acts can perform. Therefore, Dialogue act recognition is considered an important component of most spoken dialogue systems.

Many statistical models have been applied to dialogue acts classification. N-gram models can be considered the simplest method of DA classification based on some limited sequence of previous DAs as in (Hardy *et al.*, 2004; Webb, 2010; Webb and Hardy, 2005; Webb *et al.*, 2005) and sometimes used with Hidden Markova Model (HMM) as in (Boyer *et al.*, 2010). In addition, there are other approaches such as Transformation-Based Learning (TBL) as in (Samuel *et al.*, 1998), and Naïve Bayesian as in (Grau *et al.*, 2004).

Most of the previous researchers on dialogue acts classification addressed two types of feature: (1) Sentential features reflecting the linguistic characteristics of the surface utterance, which are extracted by a linguistic analyzer, such as a morphological analyzer, syntactic parser or semantic analyzer. (2) Contextual features reflecting the relationship between the current utterance and the previous utterance. In an actual dialogue, a speaker can express an identical meaning using different surface utterances based on the speaker's personal linguistic background. For this reason, it is impossible to directly compute the sentential probability because sentences are too various to find identical surface forms. To overcome this problem, researchers assume that a syntactic pattern generalizes these surface utterances using syntactic features to represent the sentential features such as sentence type, main verbs, auxiliary verbs and clue words (Choi *et al.*, 2005).

Kang et al (Kang *et al.*, 2013) proposed a model for classification speech acts for Koran language based on two-layer hierarchal structure using binary Support Vector Machine (SVM) classifiers They used the sentential features that are composed of words annotated with POS tags and POS bi-grams of all the words in an utterance and used the speech act of the previous utterance only as a contextual feature.

The proposed approach is mainly different from a Kang et al approach in three aspects - the architecture of the hierarchal structure and the selected feature set. First, the number of SVM models within their architecture requires more processing. The Authors' approach is mainly constructed using 19 SVMs models; the tested utterances are passed through the classifiers of the first layer (3 SVMs classifiers), and finally classified into one speech act among the speech acts included in the assigned type by the classifiers of the second layer (6 SVMs for Question, 7 SVMs for Response, and 3 SVMs for Other).

The proposed approach is mainly constructed using two models, one for each layer. Therefore, we think that our approach model is faster than binary classification and it can be more an efficient dialogue act classification model in real-time systems. The second difference is the number of models in multi-classes classification "our approach" not affected with a number of dialogue acts or classes but models numbers are affected when used binary classification as in the authors' approach. The third difference has they used a limited feature set that might be suitable for Koran whereas there are many features that can be used such as the relation between the speaker's dialogue act and the utterance surface, while our feature set includes rich features consisting of sentential and contextual features. For instance, speaker name, the number of utterance words, previous category, previous speaker… etc.

In fact, there are very few efforts have addressed dialogue acts classification for Arabic. (Shala *et al.*, 2010) used Naïve Bayes and Decision Trees. (Bahou *et al.*, 2008) used utterances semantic labeling based on the frame grammar formalism. (Lhioui *et al.*, 2013) used syntactic parser context-free grammar with HHM. (Graja *et al.*, 2013) used Conditional Random Fields (CRF) to semantically label spoken Tunisian dialect turns. (Hijjawi *et al.*, 2013; Hijjawi *et al.*, 2014)used Arabic function words such as "هل" "do/does", "كيف" "How" to classify questions and non-questions utterances with Decision Tree Classifier.

The proposed approach is mainly different from the previously mentioned approaches in three aspects. First, these approaches not used the hierarchal structure to solve the classification problem. Second, we provide a feature set which differed from the feature set in these approaches. We used rich features consisting of sentential and contextual features such as speaker name, the number of utterance words, previous category, previous speaker… etc. The third difference is these approaches were designed and applied on MSA or Tunisian dialect which fully differed from Egyptian dialect.

To the best of our knowledge, there is only one published work for understanding Egyptian Arabic or Egyptian dialect proposed by (Elmadany *et al.*, 2015c). They have presented a dialogue act classifier based on chunking concepts and depending on a set of sentential and contextual features. The sentential features contain four features: Utterance-Words, Words Part-Of-Speech (POS) Tags, Speaker Name, and Utterance start a label. The contextual features contain only one feature: the previous utterance act.

## 3. Utterance Labelling Model

All Let $U_{1,n}$ denote a dialogue which consists of a sequence of *n* utterances, U$_1$, U$_2$ …U$_n$, and let $DA_{1,n}$ denote the dialogue act sequences of $U_{1,n}$. Then, the dialogue act of current *utterance* can be formally defined as:

$$DA(U_i) \approx argmax_{S_{i,j}} P(CA_{i,j}|SF_i)P(CA_{i,j}|DA(U_{i-1}))$$

$DA(U_i)$ denotes the dialogue act of the i$^{th}$ utterance ($U_i$) and $CA_{i,j}$ denotes jth candidate dialogue act of the ith utterance ($U_i$), given a dialogue including n utterances ($U_{1,n}$). Therefore, we assume that the current dialogue act $DA(U_i)$ is dependent on the sentential features set ($SF_i$) of current utterance ($U_i$) and the dialogue act $DA(U_{i-1})$ of the previous utterance ($U_{i-1}$) (Choi *et al.*, 2005).

Using the utterances meta information can help dialogue acts classification process (Kim *et al.*, 2010; Ivanovic, 2005, 2008) and know what happened before current utterance can help the classification task (Sridhara *et al.*, 2009; Eugenio *et al.*, 2010). Moreover, there is a strong relationship between the speaker's dialogue act and the surface *utterances* expressing that dialogue act (Andernach, 1996; Kang *et al.*, 2013; Choi *et al.*, 2005). For instance, the speaker utters a sentence, which most well expresses his/her intention (act) so that the hearer can easily understand what the speaker's dialogue act is. On the other hand, the speaker type Operator or Customer of the current utterance can help to determine the act of utterance. For instance, the act "Service-Question" is related to the customer because he connected to service support service to asking about a provided service, but the act "Other-Question" and "Choice-Question" are related to operator because the operator asking the client about his name or choosing the client to select one of the provided services. Therefore, the sentential features represent the relationship between the dialogue acts and the surface *utterances*. In a real dialogue, the speaker utters identical contents with various surface utterances according to his personal linguistic knowledge. In addition, knowing the previous utterances acts sequence in the dialogue help the classifier to predict the dialogue act of current utterance. For instance, the act "Agree" and "Disagree" is almost followed by the "Confirm-Question" act.

The first layer of the proposed model depends on seven sentential features: Utterance-Words, Utterance-length, POS, First-Verb, Is-Part-Of-Turn, Speaker Name, and Cues; and two contextual features: speaker name of the previous utterance, and dialogue act of the previous utterance.

The second layer of the proposed model depends on eight sentential features are used: the seven sentential features as the first layer plus the main category of current utterance. So, the sentential features are Utterance-Words, Utterance-length, POS, First-Verb, Is-Part-Of-Turn, Speaker Name, Cues, and the main category of current utterance. Also, it depends on three contextual features: contextual features as the first layer plus the previous main category. So, contextual features are the main category of previous utterance, speaker name of the previous utterance, and dialogue act of the previous utterance. Table 1 shows the used sentential features in the two layers with their possible values.

| Sentential Features | Values |
|---|---|
| Utterance-Words | Uni-grams and bi-grams of utterance words |
| Utterance-Length | Number of utterance words |
| POS | Sequence of words Part-Of-Speech tags |
| First-Verb | One of four types: active (a), passive (p), not applicable (na), and undefined (u) |
| Is-Part-Of-Turn | Yes, or No |
| Speaker, Previous Speaker | Operator or Customer |
| Cue-Word and Cue-Phrase | yes, no, ok, Thank you, etc. (total of 241) |
| − The main category of current utterance <br> − The main category of the previous utterance | One of six main categories: Dialogue Structure, Turn Management, Social Obligation, Question, Answer, Social Courtesy, or Argumentation |

Table 1. Sentential Features

In the first layer, the class feature is excluded from sentential features because that is what a need to classify for is. In the second layer, the predicted class will add to sentential features. On another hand, the output of the training phase (i.e. the classification model) is used in the prediction phase to generate the final utterance act classification. In this study, WEKA[1] (Hall *et al.*, 2009), a comprehensive workbench with support for a large number of machine learning algorithms, is utilized as the development environment of the machine learning based component. The SVM algorithm is applied using SMO.

## 4. Empirical Evaluation

### 4.1 Dialogues Corpus for Egyptian Dialect

We used a corpus of real spoken dialogue in the Egyptian dialect which used in (Elmadany *et al.*, 2015c), this corpus is called JANA. JANA is a multi-genre corpus of Arabic dialogues labeled for Arabic Dialogues Language Understanding (ADLU) at utterance level and comprising Spontaneous Speech Dialogues (SSD) and Instance Messages (IM) for Egyptian dialect (Elmadany et al., 2016).

SSD has been recorded since August 2013, and it contains 52 phone calls recorded from Egyptian's banks and Egypt Air Company call-centers with an average duration of two hours of talking time after removing ads from calls. It consists of human-human discussions about providing services e.g. Create new bank account, service request, balance check and flight reservation. IM dialogues contain 30 chat dialogues, collected from mobile network operator's online-support. JANA consists of approximately 3001 turns with average 6.7 words per turn, containing 4725 utterances with average 4.3 words per utterance, and 20311 words.

---

[1] http://www.cs.waikato.ac.nz/ml/weka/

### 4.2 Experimental Results

Experimental results are presented across five datasets: Banks dataset, Flights dataset, IM dialogues dataset, combined spoken dataset (Banks and Flights), and combined dataset (Banks, Fights, and IM). Three different functions (or classifiers) are applied separately to each dataset, including SVM classifier which is supported in WEKA toolkit via SMO and built-in classifier.

Our preliminary experimental results showed that one-vs-one approach achieves the best performance in this task. Therefore, we used one-*vs*-one classification approach and the predicted probabilities are coupled using Hastie and Tibshirani's pairwise coupling method (Hastie and Tibshirani, 1998).

In this study, the evaluation is conducted based on a 10-fold cross-validation method to avoid over-fitting in which the available data set is divided into 10 folds and for each fold, a classifier is induced. The classifier is derived from 9 folds and tested on the remaining fold. The WEKA tool provides the functionality of applying the conventional k-fold cross-validation for evaluation with each classifier and then having the results represented in the aforementioned standard measures.

The first layer is classified the main category of current utterance. In the second layer, we added the classified main categories of the current and previous utterances to feature set for recognizing the dialogue act of current utterance. To test the performance of hierarchical structure in dialogue act classification and due to the lack of published works in dialect acts classification on Egyptian dialect over spontaneous dialogues either spoken or instant messages. Table 2 are illustrated the results of the proposed systems performances in terms of average F-measure when applied on Bank, Flights, IM, Combined Spoken (Banks, Flights), Combined (Banks, Flights, IM) Datasets.

| Banks | Flights | IM | Spoken | Combined Dataset |
|---|---|---|---|---|
| 0.913 | 0.902 | 0.909 | 0.909 | 0.912 |

Table 2. The results of applying our system on Bank, Flights, IM, Combined Spoken, Combined Datasets

According to the empirical results illustrated in Table 2, the overall experimental results show that the spoken dialogues highest performance than instant messages dialogues over all classifiers, and the results are much closed when applied our system using the three classifiers.

So, the results show the highest performance in acts such as Turn-Assign, Agree, SelfIntroduce, Greeting, Service-Answer, and Inform. The results show very good performance in acts such as Disagree, Service-Question, and Confirm-Question. The results show good results in acts such as Suggest and low performance in Promise, Offer, and Correct acts. The low performance due to the low counts or not exist in the training for these acts. For instance, acts 'Closing', 'Promise', and 'Offer' is not existence (i.e. N/A) in collected IM dialogues and Promise, Offer, and Correct acts are rarely existing.

In the hierarchal method, if the first layer would incorrectly classify the main category, the second layer will be classified incorrectly. For example, if the first layer is classified the main category of the current utterance as "Social Obligation", then in the second layer must choose one of four acts "Apology, Greeting, SelfIntroduce, Thanking". To solve these problems, we used the results of the first layer "main category of the current utterance and main category of the previous utterance" as features in the second layer to choose dialogue act from the 26 acts.

In Arabic dialect, especially in Egyptian Arabic, there are some words/phrases can be used in many situations with a different meaning. For example, if the operator asks "أي استفسار تاني حضرتك؟" "any other service sir?", The customer can answer "شكرا". The word "شكرا" here means "there is no other service is need" and that refers to "NO" disagree act but actually the word "شكرا" refers to thanks but here based on the dialogue it refers to NO. Also, the word "نعم" refers to "YES" agree act but sometimes used as misunderstanding sign. So, used features have solved these problems. So, the experimental results show that our system overcomes the ambiguation problem due to using the dialogue structure features such as previous act, speaker, and main category. The proposed system gives 0.909 for 'Thanking' acts, and 0.876 for 'Disagree' act in the term of F-measure. The most failure of our system due to either the rare existence or low counts of some acts in the training datasets, or there are some utterances needs to deeply semantic analysis. For instance, if the operator's utterance such as "ولكن طبعا لازم يكون عدي عليها 6 شهور" (Make sure you must get it since 6 months) and the customer has responded such as "لا لا هي عدي عليها 4 سنين" (No No it since 4 years). The system classifies the customer utterance as 'Disagree' act because it contains "لا" (No) in spite of the customer has agreed on the operator warning.

| Two-Layer Hierarchical Structure based on Binary Classification | Two-Layer Hierarchical Structure based on Multi-Classification | |
|---|---|---|
| 43.28 second | One-vs-One | One-vs-All |
| | 19.89 second | 22.94 second |

Table 3. The comparison between Two-Layer Hierarchical Structure based on Binary Classification and Multi-Classification when our system (using SMO classifier) is applied on Combined Datasets

(Kang *et al.*, 2013) has been approved using a two-layer hierarchical structure based on binary classification to solve dialogue act classification is much faster than binary classification and reported it needs only about 40% of running time of the binary classification model. The experiments results verify that the running time of two-layer hierarchical structure based on multi-classification in the training phase is much faster than a two-layer hierarchical structure based on binary classification.

Table 3 shows the comparison between Two-Layer Hierarchical Structure based on Binary Classification and Multi-Classification when our system (using SMO classifier) is applied on Combined Datasets.

| Training Models | Test datasets (Macro F-Measure) | | |
|---|---|---|---|
| | Banks | Flights | IM |
| Banks Dataset | -- | 0.855 | 0.786 |
| Flight Dataset | 0.857 | -- | 0.782 |
| IM Dataset | 0.762 | 0.778 | -- |
| All Datasets (70% train, 30% test) | 0.891 | 0.864 | 0.864 |

Table 4. The comparison results of applying proposed model on Flights and IM datasets, and Banks when using each dataset as training model

Finally, to test the generality of the proposed model on inquiry-answer domains, we trained the system using a corpus from one domain and tested the system using a corpus from a different domain. Table 4 shows the results of applying our system in the term of Macro F-measure on Flights and IM datasets, and Banks dataset when used each dataset as a training model. The results achieved the highest performance when Banks dataset has used to train the system and testing the system using others datasets 'unseen' (Flights and IM datasets). It is worth noting that we can achieve surprisingly good classification accuracy using this method.

To compare the results obtained using the proposed model with others, previous speech act analysis models in Arabic dialogues. Table 5 shows these others, previous models of different types, and their performance. We report the performance of each model as they reported and an evaluation metric that is used in their papers. So, we notice that using a hierarchical structure in dialogue acts classification has proved it's comparatively higher efficiency and improved the previous system (Elmadany et al., 2015c) results in more than 20% in the term of F-measure using same experimental setup and data.

| Classification model | Data Type | Feature set | Measurement | Score |
|---|---|---|---|---|
| (Bahou et al., 2008) | – Speech<br>– Tunisian national railway<br>– MSA | – Normalization<br>– Morphological analysis<br>– Semantic Analysis<br>– Lexical<br>– Semantic frames of the utterance. | F-Measure | 0.7179 |
| (Shala et al., 2010) | – Speech<br>– Newspaper & TV<br>– MSA | – Initial words in the utterance<br>– Parts-of-Speech<br>– Named Entity Recognition<br>– SVM, NB & J48 | F-Measure | 0.4173 |
| (Lhioui et al., 2013) | – Speech<br>– Tunisian Dialect | – Context-free grammar augmented with probabilities associated with rules<br>– HMM for creating the stochastic model | F-Measure | 0.7379 |
| (Graja et al., 2013) | – Speech<br>– Tunisian national railway<br>– Tunisian Dialect | – Conditional Random Fields (CRF)<br>– lexical normalization<br>– Morphological analysis and lemmatization<br>– Annotate word by word | F-Measure | 0.8652 |
| (Hijjawi et al., 2013)<br>(Hijjawi et al., 2014) | – Instant Messages<br>– MSA | – Arabic function words<br>– focused on classifying questions and non-questions utterances<br>– NB & Decision Tree | Accuracy | 0.8741 |
| (Neifar et al., 2014) | – Speech<br>– Tunisian national railway<br>– Tunisian Dialect | – Based on (Bahou et al., 2008)<br>– Lexical database<br>– Conceptual segmentation | F-Measure | Dataset A = 0.7322<br>Dataset B = 0.9298 |
| (Dbabis et al., 2015) | – Speech<br>– TV Programs<br>– Dialect | – Lexical<br>– Morphological<br>– Discursive and structural features<br>– SVM, NB, and J48 | F-Measure | 0.522 |
| (Graja et al., 2015) | – Speech<br>– Tunisian national railway<br>– Tunisian Dialect | – improved their previous model (Graja et al., 2013)<br>– Adding a new lexicon of the domain (Railways inquiry domain-based ontology). | F-Measure | 0.8845 |
| **(Elmadany et al., 2015c)** | – **Egyptian Dialect Dialogues (JANA corpus)** | – **Chunking concepts - Utterance-Words**<br>– **Words - Part-Of-Speech (POS) Tags - Speaker Name**<br>– **Utterance start a label -Previous utterance act** | **F-Measure** | **0.7036** |

Table 5. Performance of the proposed model and other previous models

## 5. Conclusion

This paper proposes an effective dialogue acts classification model using a multi-classes hierarchical model based on the two-layer hierarchical structure for an understanding of the Arabic dialogues task for Egyptian dialect at the utterance level. The proposed classifier has been tested using a corpus consisting of spontaneous speech dialogues and IM for Egyptian dialect, and the obtained results are very promising. In the future work, a plan is recommended to improve the classifier by adding general cues for the call-centers domain, morphological features, and dialect words treatments. Moreover, we would like to enrich the corpus with inquiry-answer dialogues from other domains e.g. Online Markets, and Railway Networks to cover 1000 Arabic dialogues.